\documentclass[dvipdfmx]{article}
\usepackage{arxiv}
\usepackage{graphicx}
\usepackage{subfigure}
\usepackage{algpseudocode}
\usepackage{algorithm}
\usepackage{amsmath}
\usepackage{newtxtext,newtxmath}
\usepackage{color}
\usepackage{bm}
\usepackage{url}
\usepackage{enumitem}
\usepackage{multicol}

\title{A Tiny Supervised ODL Core with Auto Data Pruning for Human
  Activity Recognition}

\author{
  Hiroki Matsutani\\
  Keio University\\
  3-14-1 Hiyoshi, Kohoku-ku, Yokohama, Japan\\
  \texttt{matutani@arc.ics.keio.ac.jp} \\
  \And
  Radu Marculescu\\
  The University of Texas at Austin\\
  2501 Speedway, Austin, Texas, USA\\
  \texttt{radum@utexas.edu} \\
}

\begin{document}

\maketitle
\begin{abstract}
In this paper, we introduce a low-cost and low-power tiny supervised
on-device learning (ODL) core that can address the distributional
shift of input data for human activity recognition.
Although ODL for resource-limited edge devices has been studied
recently, how exactly to provide the training labels to these devices
at runtime remains an open-issue.
To address this problem, we propose to combine an automatic data
pruning with supervised ODL to reduce the number queries needed to
acquire predicted labels from a nearby teacher device and thus save power
consumption during model retraining.
The data pruning threshold is automatically tuned, eliminating a manual 
threshold tuning.
As a tinyML solution at a few mW for the human activity recognition,
we design a supervised ODL core that supports our automatic data pruning
using a 45nm CMOS process technology.
We show that the required memory size for the core is smaller than the
same-shaped multilayer perceptron (MLP) and the power consumption is
only 3.39mW.
Experiments using a human activity recognition dataset show that the
proposed automatic data pruning reduces the communication volume by 55.7\% and
power consumption accordingly with only 0.9\% accuracy loss.
\end{abstract}

\section{Introduction}\label{sec:intro}

\begin{figure*}[t]
	\includegraphics[height=30mm]{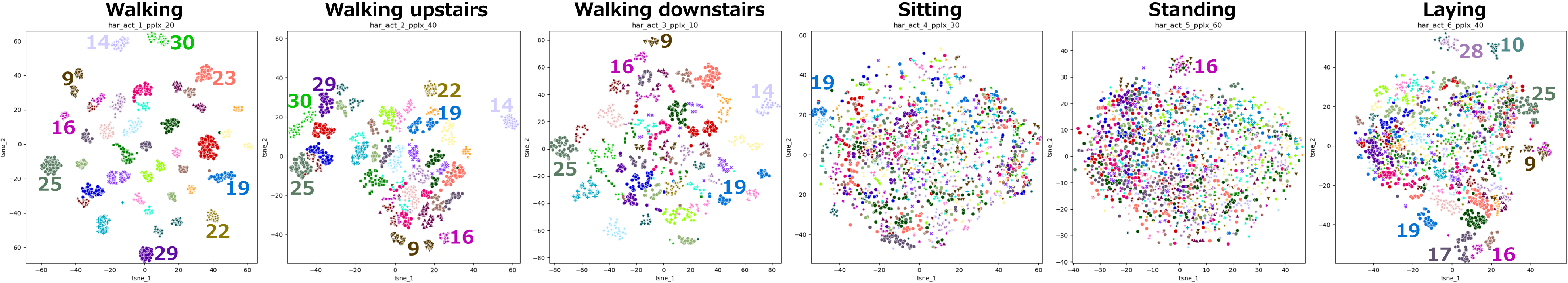}
	\caption{
2-D visualization results of 6-class human activity recognition
dataset that contains samples obtained from 30 human subjects
\cite{Reyes12}.
        }
	\label{fig:cluster}
\end{figure*}

%
%
In practical edge AI scenarios, data distribution around edge devices
may vary depending on a given environment.
Such data distribution shift is known as {\it data drift} and is a major
challenge for edge AI.
It is likely that each edge device cannot access all the data
and hence data distribution is biased depending on the environment.
In the case of human activity recognition, data distribution varies
depending on human subjects.
Figure \ref{fig:cluster} shows dimensionality reduction results of a
human activity recognition dataset that contains samples obtained from
30 human subjects \cite{Reyes12}.
There are six classes in the dataset: {\it Walking}, {\it Walking upstairs},
{\it Walking downstairs}, {\it Sitting}, {\it Standing}, and {\it Laying}.
Samples obtained from the same human subject are plotted with the same
color, while numbers in the graphs are the IDs of some selected human
subjects.
As shown in the leftmost graph, the samples from the same human
subjects form clusters during {\it Walking}.
A similar tendency is observed for the {\it Walking upstairs}, 
{\it Walking downstairs}, and {\it Laying} classes.
As mentioned above, it is likely that an edge device cannot access
the samples from all the human subjects at the design time.
In this case, an edge AI model that has been optimized for a specific human
subject may not work well for different human subjects that have not
been considered yet.

%
%
In this paper, we introduce a low-cost and low-power tiny supervised
ODL core to address the above-mentioned data drift issue for the human
activity recognition.
Although neural network ODLs for resource-limited edge devices have
been studied recently \cite{Sunaga23,Cai20,Ren21,Nadalini23}, how
exactly to provide the training labels to the edge devices at runtime for
classification tasks remains an open-issue.
For example, unsupervised or semi-supervised anomaly detection is
simply assumed in \cite{Sunaga23}.
The authors of \cite{Ren21} only mention that they enable
the onboard online post-training for 2,000 iterations via Bluetooth.
Human intervention for labeling may impose a scalability issue in the
edge AI scenarios.
To address the labeling issue, the edge devices can query a nearby teacher
device (e.g., mobile computer) wirelessly to acquire labels and
retrain the model.
However, the problem with this approach is the redundant queries to
the teacher that can waste wireless communication power.
In this paper, we propose a new approach to reduce the redundant wireless 
communication of the supervised ODL between the teacher and edge devices.

\section{Supervised ODL with Auto Pruning}\label{sec:design}

\begin{figure*}[t]
	\centering
	\includegraphics[height=42mm]{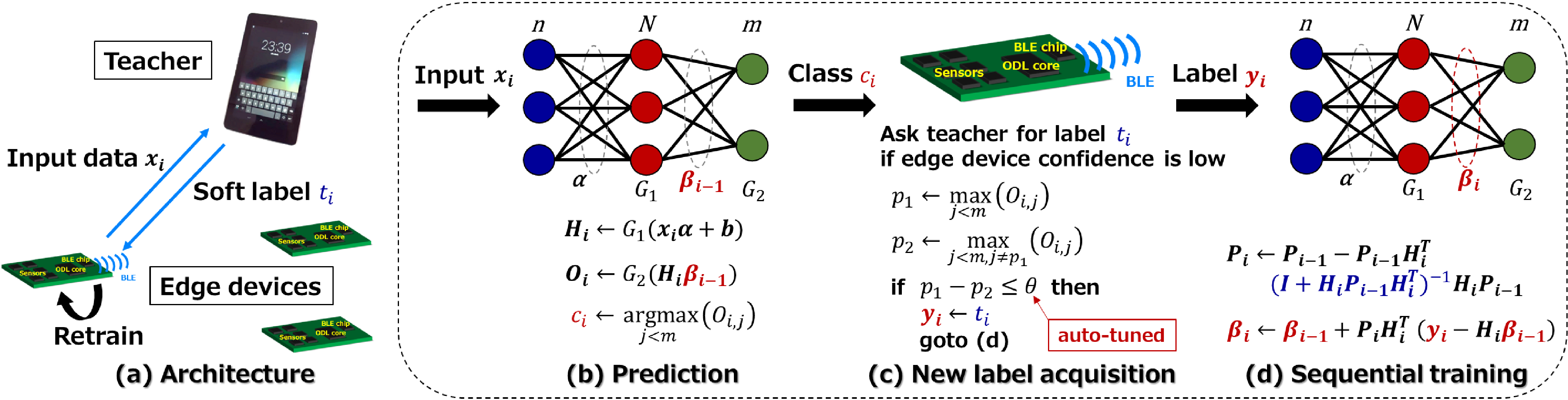}
	\caption{
Proposed supervised ODL system. 
(a) illustrates the overall system consisting of a single teacher and
multiple edge devices, where $\bm{x_i} \in \mathbb{R}^n$ and $t_i$ 
are $n$-dimensional input data and corresponding predicted label from teacher
at time $i$.
(b) illustrates the prediction algorithm at edge devices, where $n$, $N$,
and $m$ are numbers of input, hidden, and output layer nodes.
$\bm{\alpha} \in \mathbb{R}^{n \times N}$ denote weight parameters 
between input and hidden layers, and 
$\bm{\beta_{i-1}} \in \mathbb{R}^{N \times m}$ denote those between 
hidden and output layers trained at time $i-1$.
$G_1$ and $G_2$ are activation functions of hidden and output layers.
$\bm{H_i} \in \mathbb{R}^N$ and $\bm{O_i} \in \mathbb{R}^m$ are their 
outputs at time $i$.
Particularly, $O_{i,j}$ denotes probability of $j$-th class, and
$c_i$ is predicted class at time $i$.
(c) illustrates our label acquisition algorithm, where $p_1$ and $p_2$
denote probabilities of top2 labels, and $\bm{y_i} \in \mathbb{R}^m$ is 
one-hot encoded label at time $i$.
$\theta$ can be auto-tuned.
(d) illustrates the sequential training algorithm, where 
$\bm{\beta_i} \in \mathbb{R}^{N \times m}$
denote new weights updated at time $i$.
$\bm{P_i} \in \mathbb{R}^{N \times N}$ denote temporary values needed to 
compute $\bm{\beta_i}$.
        }
	\label{fig:algo}
\end{figure*}

%
%
Figure \ref{fig:algo}(a) illustrates the overall system consisting of
a single teacher and multiple edge devices.
Each edge device incorporates the proposed supervised ODL core.
We assume that the teacher is a mobile computer that has an ensemble of 
highly accurate models.
Edge devices send the input data to the teacher
and receive a predicted label from the teacher.

%
%
Algorithm \ref{alg:top} shows the top-level function of the edge
devices;
this function is called periodically (e.g., once per second) by the
edge devices.
Assume that one of the edge devices in Figure \ref{fig:algo}(a)
collects some sensor data $\bm{x_i}$ at time $i$ (Line 1).
The next step depends on the operation mode of the ODL core: predicting
or training mode.
An ODL algorithm based on neural networks is used to update their weight
parameters in the training mode.
Section \ref{ssec:odl_algo} describes the ODL algorithm (Lines 6 and 9).
Section \ref{ssec:label} describes our label acquisition approach (Line 8) 
via the teacher device and automatic data pruning to reduce communication.
The operation modes are switched by a data drift detection
algorithm (Line 3).
Existing data drift detection algorithms \cite{Yamada23}
can be used by considering expected data drift types such as sudden
drift, gradual drift, incremental drift, and reoccurring drift.
The training finishes when a prespecified condition, such as the
number of required training samples or training loss, is satisfied
(Line 10).
Finally, Section \ref{ssec:core} describes the design of
the proposed tiny supervised ODL core.

\begin{algorithm}[t]
\caption{Top-level function (at time $i$) of our ODL core.}
\label{alg:top}
\begin{algorithmic}[1]
\State $\bm{x_i} \gets$ Sense()
\If {mode = predicting} 
	\If {IsDrift($\bm{x_i}$)}
		\State mode $\gets$ training
	\EndIf
	\State \Return Predict($\bm{x_i}$) \Comment{Figure 2(b)}
\ElsIf {mode = training}
	\State $\bm{y_i} \gets$ LabelAcquire(Predict($\bm{x_i}$)) \Comment{Figure 2(c)}
	\State SequentialTrain($\bm{x_i}$, $\bm{y_i}$) \Comment{Figure 2(d)}
	\If {IsTrainDone()}
		\State mode $\gets$ predicting
	\EndIf
\EndIf
\end{algorithmic}
\end{algorithm}

\subsection{ODL Algorithm}\label{ssec:odl_algo}

%
%
To enable ODL on resource-limited edge devices, we utilize the OS-ELM
algorithm in \cite{Liang06}.
OS-ELM assumes simple neural networks that consist of input, hidden,
and output layers, as shown in Figure \ref{fig:algo}(b).
The weight parameters between the input and hidden layers are denoted
as $\bm{\alpha}$, while those between the hidden and output layers are
denoted as $\bm{\beta}$.
$\bm{\alpha}$ are initialized with random values at the startup time,
while $\bm{\beta}$ are sequentially updated by input data $\bm{x}$ and
labels $\bm{y}$.
%
%
%
%
Figure \ref{fig:algo}(d) shows the sequential training algorithm of OS-ELM.
At time $i$, $\bm{\beta_i}$ are calculated and updated by $\bm{x_i}$,
$\bm{y_i}$, previous weight parameters $\bm{\beta_{i-1}}$, and 
temporary values $\bm{P_i}$.
As shown in the figure, $\bm{P_i}$ are derived from the previous
values $\bm{P_{i-1}}$ and $\bm{H_i}$, which are hidden-layer outputs
based on $\bm{x_i}$.

\subsection{Label Acquisition with Auto Data Pruning}\label{ssec:label}

%
%
Since proper labeling is mandatory to correctly train the model in a
short time, we exploit the nearby teacher device via BLE;
that is, the edge devices send input data $\bm{x_i}$ to the nearby 
teacher and receive a predicted label $t_i$ from the teacher, as shown
in Figure \ref{fig:algo}(c). 
$t_i$ is then converted to a one-hot encoded label $\bm{y_i}$.
If such a nearby teacher is not available, the queries to the teacher will
be retried later or skipped.
The problem with this approach is the excessive number of queries to
the teacher that can waste wireless communication power.

%
%
We propose to use data pruning
that suppresses excessive queries combined with the supervised ODL.
In our label acquisition approach, if the conditions below are met:
\begin{enumerate}
\item A pre-specified number of samples has been trained,
\item Data drift is not currently detected, and
\item Confidence of the predicted label is high,
\end{enumerate}
then the edge devices do not need to query a teacher, and thus they
can skip the sequential training.
The first and second conditions confirm that a data distribution did
not change during this training phase.
As for the third condition, 
in this paper, our edge devices predict locally using the current weights
being trained to produce the first and second highest probabilities
$p_1$ and $p_2$ of top2 labels, as shown in Figure \ref{fig:algo}(c).
Then, their difference $p_1 - p_2$ is used as a confidence metric, which 
is denoted as ``P1P2'' in this paper.
That is, the third condition is satisfied if $p_1 - p_2 > \theta$.
$\theta$ is a tuning parameter, and a manual tuning of $\theta$ by 
sweeping many possible values is {\it impractical}.
To properly choose $\theta$ at runtime,
\begin{enumerate}
\item $\theta$ is set to a high value at the startup time;
\item $\theta$ is then gradually decreased if the following condition is 
	$X$ times consecutively satisfied: $p_1 - p_2 > \theta$ or $c = t$ 
	when querying ($p_1 - p_2 \le \theta$),
	where $c$ is locally-predicted label and $t$ is teacher's predicted 
	label;
\item $\theta$ is increased if $c \neq t$ when querying 
	($p_1 - p_2 \le \theta$).
\end{enumerate}

\subsection{ODL Core Design}\label{ssec:core}


%
%
As a tinyML solution at a few mW for the human activity recognition,
the proposed ODL core including the prediction, label acquisition, and
sequential training is designed with Verilog HDL.
%
The number of input layer nodes $n$, hidden layer nodes $N$, and
output layer nodes $m$ can be customized; 
for example, they are set to 561, 128, and 6 in our prototype 
implementation. 
This ODL core is {\it flexible}.
Because the multiply-add and division units are controlled by a state
machine for the prediction, label acquisition, and sequential
training, $n$, $N$, and $m$ can be changed at the startup time
depending on applications as long as the on-chip SRAM capacity allows
it.
%

%
%

%
%
In OS-ELM, random values are stored as $\bm{\alpha}$
\cite{Liang06}.
To reduce the memory size of $\bm{\alpha}$, the weights can be
replaced with a pseudorandom number generator as in \cite{Li19}.
We also follow this direction and hence examine the following variants:
\begin{itemize}
\item ODLBase: 32-bit random numbers are stored as weights $\bm{\alpha}$.
\item ODLHash: $\bm{\alpha}$ are replaced with a 16-bit
  Xorshift function, where coefficients are 7, 9, and 8.
\end{itemize}

\begin{table}[t]
\centering
\caption{Memory size of ODL cores [kB] ($n$ = 561 and $m$ = 6).}
\label{tab:mem}
\begin{tabular}{l|rrrrr}
\hline \hline
$N$		&32	&64	&128	&256	&512\\
\hline
NoODL		&74.82	&147.40	&292.55	&582.85	&1163.46	\\
ODLBase		&83.01	&180.16	&423.62	&1107.14&3260.61	\\
ODLHash		&11.20	&36.55	&{\bf 136.39}	&532.68	&2111.68	\\
\hline
\end{tabular}
\end{table}

\begin{table}[t]
\centering
\caption{Comparisons with reported results.} 
\label{tab:cmp}
\begin{tabular}{l|rr}
\hline \hline
			&\# of parameters & Accuracy [\%] \\
\hline
ODLHash ($N$ = 128)		& 34k	& 93.67	\\
ODLHash ($N$ = 256)		& 133k	& 95.51 \\
Q. Teng et al., \cite{Teng20} 	& 0.35M	& 96.98 \\
W. Huang et al., \cite{Huang21}	& 0.84M	& 97.28 \\
\hline
\end{tabular}
\end{table}

%
%
Table \ref{tab:mem} shows their total memory sizes $M_{bit}$ when $N$
is varied from 32 to 512 assuming that $n$ and $m$ are 561 and 6,
respectively.
``NoODL'' means the same MLP as in ODLBase, but without the ODL 
capability.
As shown in the table, $M_{bit}$ is significantly reduced by 
ODLHash compared with ODLBase especially when $N$ is less than
or equal to 256.
It is interesting that in these cases ODLHash is smaller than
NoODL.

Table \ref{tab:cmp} compares ODLHash with results reported in
\cite{Teng20,Huang21} in terms of the parameter size and
classification accuracy for the human activity recognition dataset
\cite{Reyes12}.
Our ODLHash achieves favorable accuracies compared with these SOTA
results even with {\it very small} parameter sizes (e.g., ODLHash when $N$
is 128 is an order of magnitude smaller than \cite{Teng20} and \cite{Huang21}).
More importantly, ODLHash has the ODL capability (thus including 
the temporary storage for ODL) which is a big advantage when it is 
deployed at environments where data distribution may change, as 
demonstrated in the next section.


\section{Evaluations}\label{sec:eval}


%
%
The proposed ODL approaches are evaluated in terms of the
classification accuracy using a data drift dataset.
%
%
%
The human activity recognition dataset \cite{Reyes12}
is modified to create shifted datasets before and after a data drift.
Specifically, based on the dimensionality reduction results (Figure
\ref{fig:cluster}), samples obtained from human subjects 9, 14, 16,
19, and 25 are removed from the original training and test datasets,
and these reduced training and test datasets are used as training and
``test0'' datasets in this paper.
Those of human subjects 9, 14, 16, 19, and 25 are used 
as ``test1'' dataset.
Labels of these datasets are used as teacher's predicted labels 
(i.e., $t_i$ in Figure \ref{fig:algo}(c)).

\begin{table}[t]
\centering
\caption{Accuracy of ODL approaches and counterparts before and after drift.}
\label{tab:acc_hash}
\begin{tabular}{l|rr}
\hline \hline
			& Before [\%]	& After [\%]\\ \hline
NoODL ($N$ = 128)	& 92.9$\pm$0.8	& 82.9$\pm$1.4 \\
ODLBase ($N$ = 128)	& 93.4$\pm$0.6	& 90.8$\pm$1.7 \\
ODLHash ($N$ = 128)	& 93.1$\pm$0.8	& 90.7$\pm$1.0 \\
\hline
NoODL ($N$ = 256)	& 95.1$\pm$0.3	& 83.7$\pm$1.0 \\
ODLBase ($N$ = 256)	& 95.2$\pm$0.3	& 92.5$\pm$0.6 \\
ODLHash ($N$ = 256)	& 95.1$\pm$0.4	& 92.3$\pm$0.7 \\
\hline
DNN (561,512,256,6)	& 94.1$\pm$1.0	& 85.2$\pm$1.3 \\
\hline
\end{tabular}
\end{table}

\begin{figure*}[t]
\begin{minipage}[h]{0.49\linewidth}
	\centering
	\includegraphics[height=50mm]{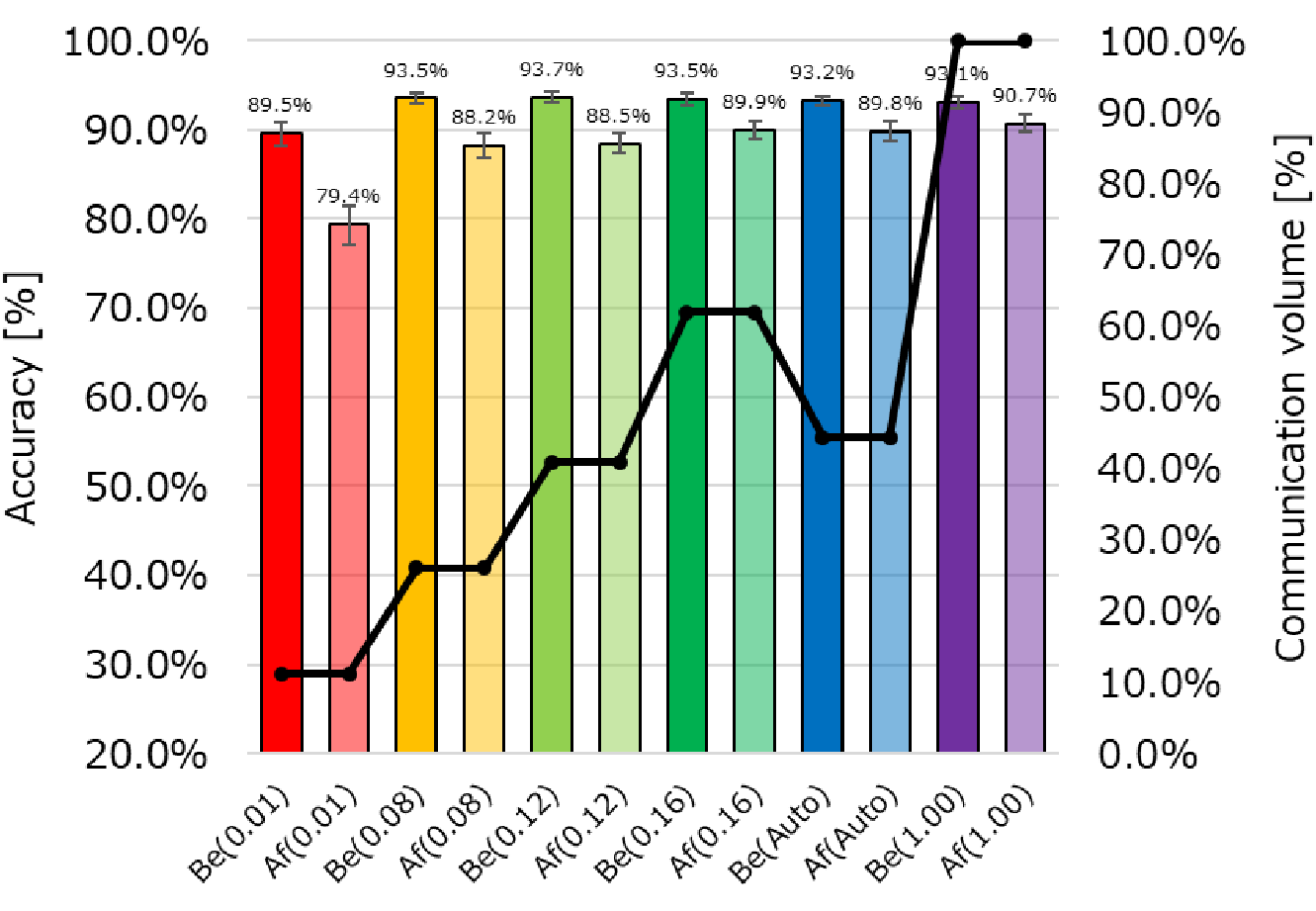}
	\caption{Communication volume with different $\theta$.}
	\label{fig:acc_conf}
\end{minipage}
\begin{minipage}[h]{0.49\linewidth}
	\centering
	\includegraphics[height=50mm]{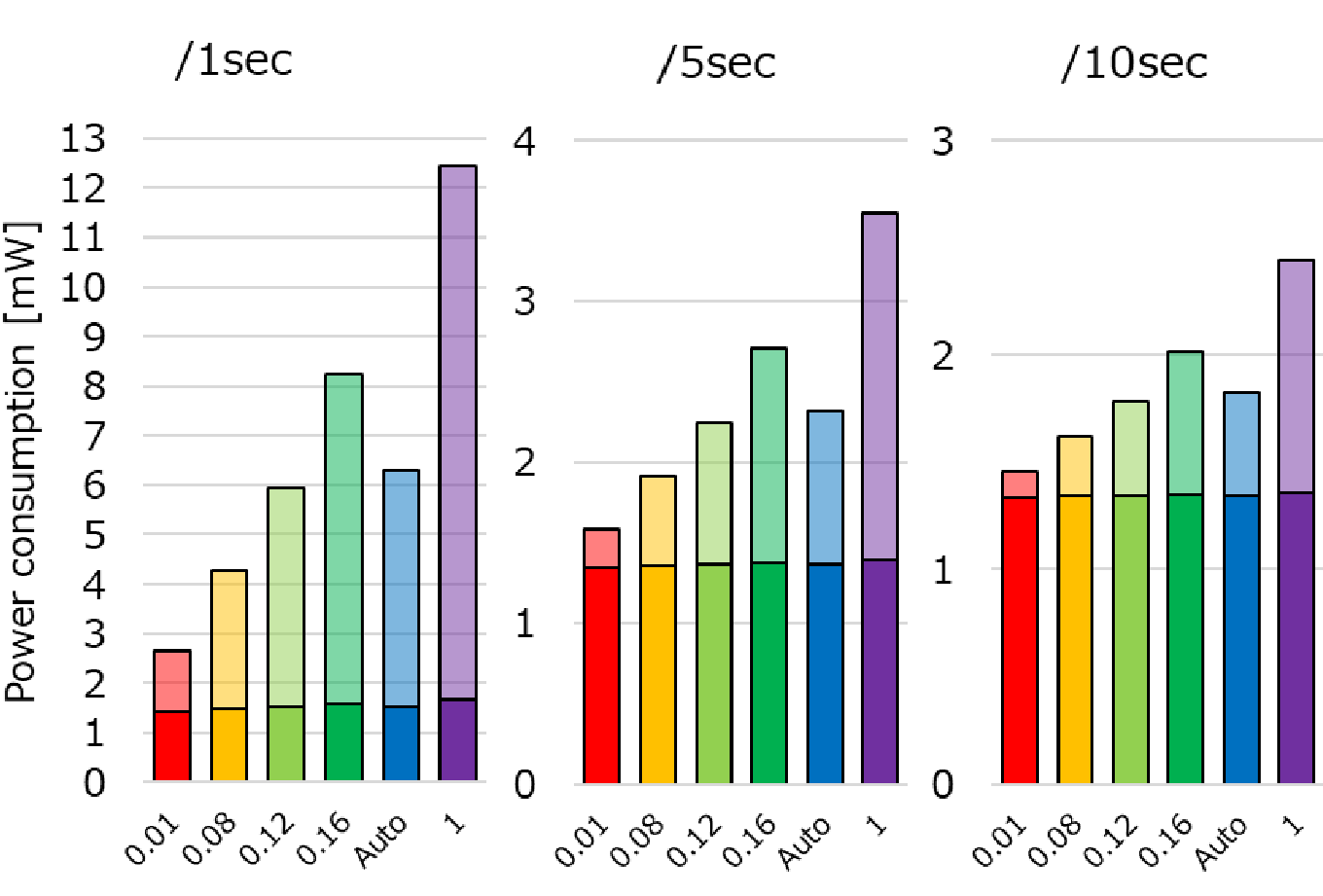}
	\caption{Power consumption with different $\theta$.}
	\label{fig:power}
\end{minipage}
\end{figure*}

%
%
The following steps are executed 20 times for each approach to
report their test accuracies before and after the data drift:
\begin{enumerate}
\item Initial training: Models are trained with the training
  dataset.
\item Test before drift: Models are tested with the test0 dataset.
\item ODL: Models are retrained with approximately 60\% of the test1
  dataset.
  This step is not executed in NoODL.
\item Test after drift: Models are tested with the remaining samples of
  the test1 dataset.
\end{enumerate}

\subsection{ODL Approaches vs. NoODL}\label{ssec:acc_hash}

%
%
The proposed ODL approaches and counterparts are evaluated in terms 
of the classification accuracy.
Table \ref{tab:acc_hash} shows the mean and standard deviation of the
test accuracies before and after the data drift.
``DNN'' is a simple MLP that has two hidden layers.
The accuracies tend to be high when $N$ is large, 
and they are almost saturated when $N$ is 256.
Although the accuracies when $N$ is 128 are slightly lower than those
when $N$ is 256, the memory sizes when $N$ is 256 are significantly
large compared with those when $N$ is 128; for example, the memory
size of ODLHash is increased by 3.91$\times$ as $N$ is increased from 128 to
256, as shown in Table \ref{tab:mem}.
We thus focus on ODL approaches when $N$ is 128 since they can strike
a good balance between the accuracy and memory size as low-end tinyML
solutions.
%
%
In NoODL, the accuracy drops after the data drift, while the accuracy
is recovered in ODLBase and ODLHash.
ODLHash achieves almost the same accuracies as ODLBase.
These results demonstrate benefits of our supervised ODL in data
drift situations.
%
%

\subsection{Data Pruning with Different Thresholds}\label{ssec:acc_conf}

%
%
The proposed ODLHash with the data pruning (P1P2) is evaluated in 
terms of the classification accuracy and
communication volume by varying the confidence threshold $\theta$.
$N$ is set to 128.
$\theta$ is varied from 0.01 to 1 (no data pruning when $\theta$ = 1).
The experiments are executed 20 times for each configuration.
The number of initial samples to be trained before the data pruning is
enabled (i.e., condition 1 of Section \ref{ssec:label}) is empirically
determined by max($N$, 288).

%
%
In Figure \ref{fig:acc_conf}, bar graphs show the mean and standard
deviation of the test accuracies before and after the data drift.
``Be'' and ``Af'' indicate those of ODLHash before and after the data drift.
``Auto'' indicates the proposed auto-tuning of $\theta$.
As expected, the accuracies tend to be low when $\theta$ is 0.01.
However, the accuracy drop is not very large when $\theta \geq$ 0.08.
%
%
In Figure \ref{fig:acc_conf}, the line graph shows the communication
volume between edge and teacher devices.
In our approach, the communication occurs during the training mode.
The communication volume without the data pruning is assumed as 100\%.
Overall, the volume is significantly reduced by the data
pruning in exchange for a negligible or small accuracy loss.
This is because the dataset contains a lot of similar samples and thus
the data redundancy is high.
Such situations can be easily expected when continuously-generated
sensor data streams are learned by edge devices.
Finally, Be(Auto) and Af(Auto) in Figure \ref{fig:acc_conf} show the
accuracies when $\theta$ is auto-tuned broadly from among 1, 0.64,
0.32, 0.16, and 0.08.
$X$ is set to 10, which is a conservative configuration.
The accuracy is decreased by up to 0.9\%, while the communication
volume is decreased by 55.7\% compared to that when $\theta$ is 1.
Comparisons to the other data pruning metrics (e.g., Error L2-Norm 
\cite{Paul21}) are omitted due to page limitation.

\subsection{Power Savings: A Case Study}\label{ssec:power}

\begin{figure}[t]
\begin{minipage}{0.29\linewidth}
	\centering
	\includegraphics[height=35mm]{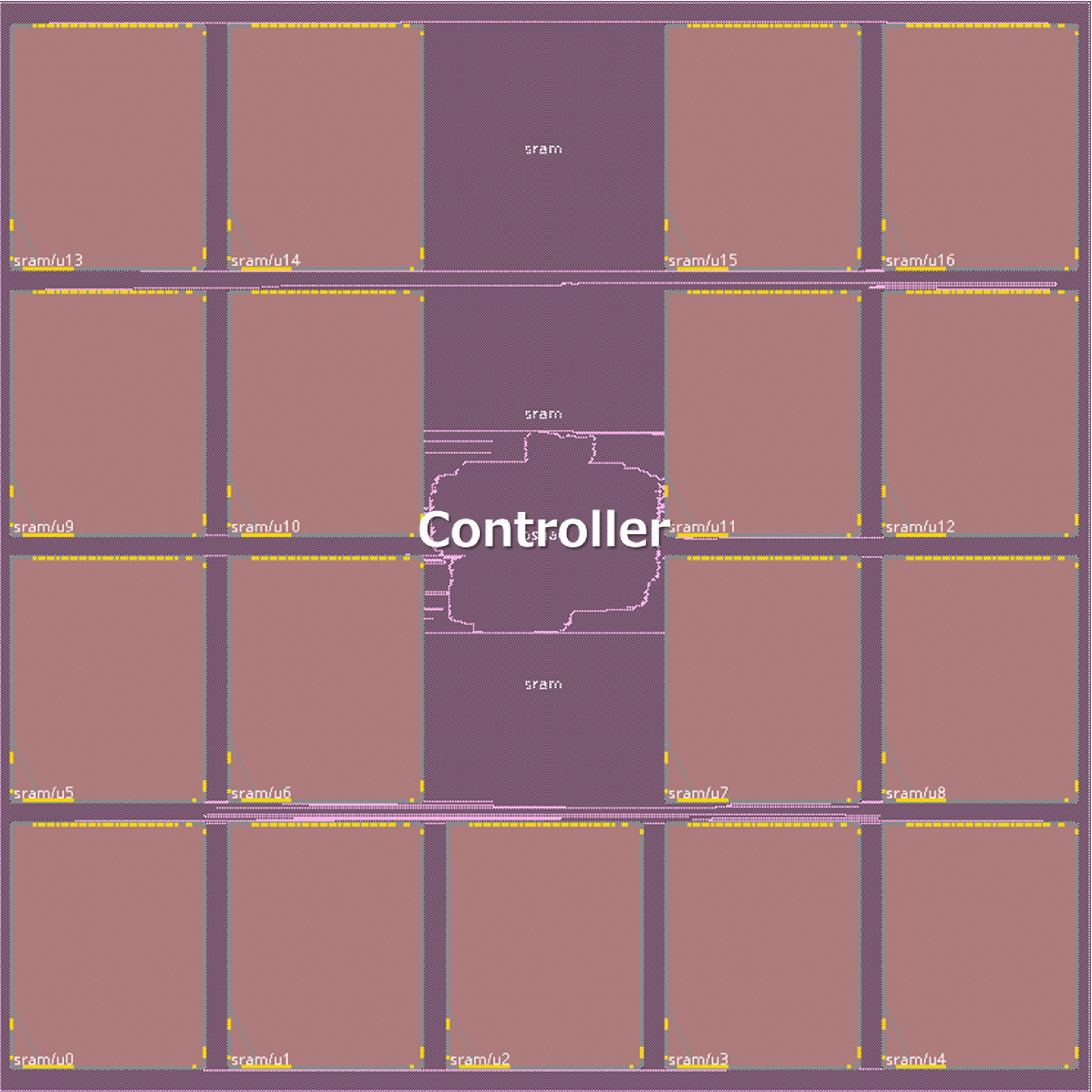}
	\caption{ODL core layout.}
	\label{fig:layout}
\end{minipage}
\begin{minipage}{0.70\linewidth}
	\centering
	\makeatletter
	\def\@captype{table}
	\makeatother
	\caption{Execution time and power consumption of ODL core at 10MHz.}
	\label{tab:core}
	\begin{tabular}{l|r}
	\hline \hline
	Core size        & 2.25mm $\times$ 2.25mm \\ \hline
	Prediction time  &  36.40 [msec] \\
	Seq. train time  & 171.28 [msec] \\ \hline
	Prediction power &   3.39 [mW] \\ 
	Seq. train power &   3.37 [mW] \\ 
	Idle power       &   3.06 [mW] \\ 
	Sleep power      &   1.33 [mW] \\ 
	\hline
	\end{tabular}
\end{minipage}
\end{figure}

%
%
The communication volume reduction by the
proposed data pruning is evaluated in terms of power savings.
%
%
ODLHash where $n$, $N$, and $m$ are 561, 128, and 6, respectively, is
implemented with the Nangate 45nm Open Cell Library.
Numbers are represented by 32-bit fixed-point format.
The memory size is only 136.39kB, and it is implemented with 17 8kB SRAM
cells (see the layout in Figure \ref{fig:layout}).
%
%
Table \ref{tab:core} shows the execution times of
prediction and sequential training when the operating frequency is 10MHz.
Even with this very low frequency, the sequential training time is
171msec, which is fast enough for a per-second operation; thus, the
frequency is set to 10MHz in this paper.
Table \ref{tab:core} also shows the power consumptions of
prediction and sequential training when the frequency is 10MHz.
They are derived from switching activities of the logic and memory parts
observed in post-layout simulations.
We note that the logic part is stateless and can be powered off
when it is not used, while the memory part cannot since it is
retaining weights and states.
The sleep mode power in Table \ref{tab:core} is estimated with this assumption.

%
%
The power consumption during the training mode contains those for
communication (label acquisition) and computation (prediction and 
sequential training).
Regarding the communication power, we assume that edge devices use BLE
to send 561 features to a teacher device and receive the corresponding
label from the teacher.
We assume Nordic Semiconductor nRF52840 as a BLE chip.
Data rate is 1Mbps, TX power is 0dBm, and supply voltage is 3.0V.
The power values are estimated by Nordic Semiconductor online tool
\cite{NordicPower}.

%
%
Figure \ref{fig:power} shows the total power consumption of the
ODLHash core during the training mode with the data pruning by varying
$\theta$.
$N$ is set to 128.
In these bar graphs, the dark parts represent computation power, while 
the light parts represent communication power.
The results highly depend on the frequency of operations.
A series of sensing, prediction, and sequential training operations is
called ``event'' here.
We assume three event frequencies: one event per second, one event per
5 seconds, and one event per 10 seconds.
In the one event per second case, the power consumption is reduced by
33.8\%-65.7\% when $\theta$ ranges between 0.16 and 0.08; 
it is reduced by 23.7\%-46.1\% in the once per 5 seconds case, and
by 17.3\%-33.5\% in the once per 10 seconds case.
Finally, ``Auto'' in Figure \ref{fig:power} shows the power
consumption when $\theta$ is auto-tuned conservatively as mentioned in
Section \ref{ssec:acc_conf}.
The power consumption is reduced by 49.4\%, 34.7\%, and 25.2\% in the
once per second, once per 5 seconds, and once per 10 seconds cases.
The accuracy drop is only 0.9\%.
A smaller $X$ saves more power while it affects the accuracy.


\section{Conclusions}\label{sec:conc}

The proposed tiny supervised ODL core (ODLHash ($N$ = 128)) that 
supports the automatic data pruning runs at only 3.39mW. 
The memory size is only 136.39kB.
Our experiments using the drifted dataset show that although the
proposed ODLHash is smaller than the NoODL baseline, it can recover
accuracy by ODL when data drift occurs.
The results also show that our automatic data pruning reduces the
communication volume by 55.7\% and training mode power on the tiny
supervised ODL core significantly with a small accuracy loss.
%


{\bf Acknowledgements~~}
H.M. was supported in part by JST AIP Acceleration Research JPMJCR23U3, Japan.
H.M. acknowledges supports from VLSI Design and Education Center (VDEC).


\end{document}